\documentclass[letterpaper, 10 pt, conference]{ieeeconf}  

\usepackage{xcolor,colortbl}
\usepackage{ucs}
\usepackage{eso-pic}
\usepackage{graphicx}
\usepackage{epstopdf}
\usepackage{amsmath}

\usepackage{accents}

\usepackage{subfig}
\usepackage{datatool}
\usepackage{filecontents,pgfplots}
\usepackage{pgfplots}
\usepackage{mathtools}
\usepackage{pgf,tikz}

\usepackage{lmodern}
\usepackage{float}
\usepackage{longtable}
\usepackage{pdflscape}
\usepackage{multicol}
\usepackage{graphicx}
\usepackage{epstopdf}
\usepackage{url}

\usepackage{amssymb,amsmath,amstext,amsthm}
\usepackage{enumerate}
\usepackage{listings}
\usepackage{subfig}
\usepackage{datatool}
\usepackage{filecontents,pgfplots}
\usepackage{pgfplots}

\usepackage{tikz}
\usepackage[lcgreekalpha]{stix}

\usetikzlibrary{decorations.pathmorphing} 
\usetikzlibrary{fit}					
\usetikzlibrary{backgrounds}	

\IEEEoverridecommandlockouts                              

\title{\LARGE \bf
	Cognitive and motor compliance in intentional human-robot interaction 
}

\author{Hendry F. Chame$^{\dagger}$ and Jun Tani$^{\dagger}$
	\thanks{${}^\dagger$ Cognitive Neurorobotics Research Unit (CNRU), Okinawa Institute of Science and Technology (OIST), Okinawa, Japan 904-0495.}
	\thanks{E-mail: \{hendryfchame,\ tani1216jp\}@gmail.com}%
}

\definecolor{plotRed}{HTML}{B22222}
\definecolor{plotBlue}{HTML}{1E90FF}
\definecolor{plotGreen}{HTML}{006400}

\begin{document}

\AddToShipoutPictureBG*{%
  \AtPageUpperLeft{%
    \hspace{19.4cm}%
    \raisebox{-1.2cm}{%
      \makebox[0pt][r]{\large\textcolor{plotGreen}{Accepted for publication in the 2020 International Conference on Robotics and Automation (ICRA)}}
}}}%

	\maketitle
	\thispagestyle{empty}
	\pagestyle{empty}

	\begin{abstract}
		
		Embodiment and subjective experience in human-robot interaction are important aspects to consider when studying both natural cognition and adaptive robotics to human environments. Although several researches have focused on nonverbal communication and collaboration, the study of autonomous physical interaction has obtained less attention. From the perspective of neurorobotics, we investigate the relation between intentionality, motor compliance, cognitive compliance, and behavior emergence. We propose a variational model inspired by the principles of predictive coding and active inference to study intentionality and cognitive compliance, and an intermittent control concept for motor deliberation and compliance based on torque feed-back. Our experiments with the humanoid Torobo portrait interesting perspectives for the bio-inspired study of developmental and social processes.

	\end{abstract}

	\section{INTRODUCTION}

	Our society is probably changing into a world populated by natural and artificial beings, where we will coexist with robots at home and the office. However, at present, there is still an important gap to be overcome, in particular, when we pause to contemplate the amazing  complexity of natural behavior. This sort of sophistication is related to important properties, among which are autonomy and intentionality.
	
	When studying motor interaction between a human and a robot, some works have focused on goal-directed collaboration (e.g. mediated by physical objects \cite{pelle2016}, and behavior improvement \cite{landi2018}), leaving aside the social dimension of the interaction with the robot partner. With some exceptions, direct contact in autonomous behavior has been avoided \cite{kollmitz2018}. Moreover, in social robotics, the non-verbal aspects of interaction have been studied from diverse modalities (e.g. facial expressions \cite{ishi2019} and touch \cite{hu2019}), though motor clues in direct contact have attracted less attention.
	
	In psychology, physical interaction is considered an intuitive means of communication during human early life, which underlies the acquisition and development of social and cognitive skills \cite{vygotsky1980}. Hence, it is paramount for human development, and it is relevant for studying learning from the perspective of developmental robotics \cite{Ikemoto12}.
	

	Intentional motor interaction is certainly a broad phenomenon that ought to be delimited. Thus, by taking inspiration on neuroscience research, we hold the assumption that intentionality involves an optimization process in hierarchical representation structures, in which a top-down information flow characterizes the agency of purposeful actions in a given context, whereas a bottom-up information flow accounts for their consequences. We investigate this within the perspective of Friston's \textit{free energy principle} theory \cite{friston2010}, according to which the brain attempts to resolve conflicts possibly appearing between the top-down information flow, developed by a generative model, and the bottom-up sensory flow, through minimizing free energy as a statistical quantity.
	
	Our research is interested in the study of harmonious (or coherent) and disharmonious (or incoherent) interaction between the human and the robot. Thus, an important distinction is established between physical (or motor) and cognitive (or representational) compliance. The former is defined as the capacity to be driven by external physical action, which can be conforming or conflicting with the intended action; whereas the later involves the flexibility in modifying the generative process according to information from the ascending process, that is, the capacity to be driven by sensory evidence (the inference process). 
	
	After having established the previous considerations, we claim that the originality and main contribution of our research is to study the relation between physical and cognitive compliance in purposeful motor interaction. Within the perspective of neurorobotics, we propose a variational model, inspired by the principles of predictive coding and active inference \cite{friston2011}, to study intentionality and cognitive compliance. We propose an intermittent control concept for the study of motor deliberation and compliance that takes into account torque feed-back. From the analysis of interaction and behavior emergence in experiments with the human-sized Torobo platform, we illustrate the relevance of our work to the study of direct interaction, with interesting perspectives for the bio-inspired approach to developmental and social robotics.  
	
	\section{RELATED WORK}
	\label{sec:relatedWork}
	
	Although independently developed, our research has been consistent with the \textit{free energy principle} theory \cite{friston2011}. In \cite{tani2003} a deterministic hierarchical neural network architecture operating on different time scales (i.e. a Multiple Timescale Recurrent Neural Network, or MTRNN) was proposed for learning temporal sequences. Inspired by these ideas, behavior imitation was studied from visual and proprioceptive representations (e.g. in \cite{hwang2018}).
	
	In order to improve generalization, stochastic modeling has been adopted where uncertainty in training data is learned as a Gaussian distribution in the output layer \cite{murata2013}. However, a limitation of this approach is the fact that the context layers in the hierarchy remained deterministic. 
	
	The emergence of the variational Bayes auto-encoder (VAE) framework  \cite{kingma2013} paved the way for optimizing inference and learning probabilistic distributions in latent variables, by re-parameterizing the variational lower bound. This framework has been extended for studying on-line interaction, as for example anticipating human behavior \cite {butepage2018}.

	
		The \textit{predictive-coding-inspired variational recurrent neural network} (PV-RNN) \cite{ahmadi2019} has been proposed recently in our lab. It is a variational model capable of learning probabilistic structures in fluctuating temporal patterns, by modifying dynamically the stochasticity of the represented latent states. Different from VAEs, inputs are not propagated in the network during forward computations. Instead, prediction errors are \textit{back propagated through time} (BPTT). 
		
	Unlike \cite{hwang2018}, where the robot imitates the human behavior from visual input, our work is focused on direct physical interaction based exclusively on the proprioceptive source of information. Since our interest is to investigate autonomous and possibly conflicting interaction, different from \cite {butepage2018}, our study does not focus on predicting the human behavior, but on the inference process over proprioceptive representations. Thus, the agent receives the human intentions through how its body posture is changed within the interaction, which characterizes a form of \textit{primary intersubjective} experience \cite{spaulding2012}, as studied in \textit{embodied social cognition} (ESC) theory. 
	
	\section{COGNITIVE COMPLIANCE}
	\label{sec:Rnn}
	
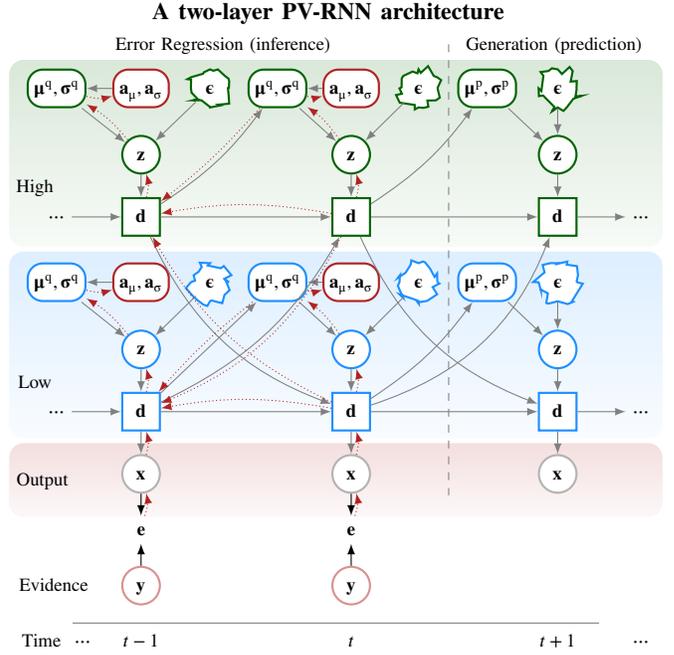
\begin{figure}[htbp]
		\centering
		\begin{scriptsize}
			\tikzstyle{state1}=[circle,
			thick,
			minimum size=0.5cm,
			draw=plotGreen,
			fill=white!40]
			
			\tikzstyle{prediction}=[circle,
			thick,
			minimum size=0.5cm,
			draw=black!30,
			fill=white!40]
			
			\tikzstyle{sensation}=[circle,
			thick,
			minimum size=0.5cm,
			draw=plotRed!50,
			fill=white!40]
			
			\tikzstyle{unitA}=[rectangle,
			thick,
			minimum size=0.5cm,
			draw=plotRed,
			fill=white!40,
			rounded corners=2mm]
			
			\tikzstyle{unitMS1}=[rectangle,
			thick,
			minimum size=0.5cm,
			draw=plotGreen,
			fill=white!40,
			rounded corners=2mm]
			
			\tikzstyle{unitD1}=[rectangle,
			thick,
			minimum size=0.5cm,
			draw=plotGreen,
			fill=white!40]
			\tikzstyle{noise1}=[circle,
			thick,
			minimum size=0.5cm,
			draw=plotGreen,
			fill=white!40,
			decorate,
			decoration={random steps,
				segment length=2pt,
				amplitude=2pt}]
			
			\tikzstyle{state0}=[circle,
			thick,
			minimum size=0.5cm,
			draw=plotBlue,
			fill=white!40]
			
			\tikzstyle{unitMS0}=[rectangle,
			thick,
			minimum size=0.5cm,
			draw=plotBlue,
			fill=white!40,
			rounded corners=2mm]
			
			\tikzstyle{unitD0}=[rectangle,
			thick,
			minimum size=0.5cm,
			draw=plotBlue,
			fill=white!40]
			\tikzstyle{noise0}=[circle,
			thick,
			minimum size=0.5cm,
			draw=plotBlue,
			fill=white!40,
			decorate,
			decoration={random steps,
				segment length=2pt,
				amplitude=2pt}]
			
			\tikzstyle{backgroundL1}=[rectangle,
			draw=black,
			fill=white!40,
			inner sep=0.2cm,
			rounded corners=2mm]
			
			\tikzstyle{backgroundL0}=[rectangle,
			draw=black,
			fill=white!40,
			inner sep=0.2cm,
			rounded corners=5mm]
			
			\tikzstyle{backgroundTime}=[rectangle,
			fill=white,
			inner sep=0.2cm,
			rounded corners=5mm]
			\begin{tikzpicture}[>=latex,text height=1.5ex,text depth=0.25ex]
			
			\node [color=black] at (-0.1,4.6) {\small \textbf{A two-layer PV-RNN architecture}};				
						

			
			\node (rect) at (0.0,2.75) [draw=white,top color=plotGreen!15,bottom color=plotGreen!3,rounded corners=2mm,minimum width=8.7cm,minimum height=2.5cm] {};
			\node (rect) at (0.0,0.2) [draw=white,top color=plotBlue!15,bottom color=plotBlue!3,rounded corners=2mm,minimum width=8.7cm,minimum height=2.5cm] {};			
			\node (rect) at (0.0,-1.6) [draw=white,top color=plotRed!15,bottom color=plotRed!3,rounded corners=2mm,minimum width=8.7cm,minimum height=1.0cm] {};
			
			\node at (-1.5,4.2) {Error Regression (inference)};
			\node at (2.9,4.2) {Generation (prediction)};
			
			\node at (-4,2.3) {High};
			\node at (-4,-0.3) {Low};
			\node at (-3.9,-1.6) {Output};
			\node at (-3.75,-3.0) {Evidence};

			\draw[gray,dashed] (1.5,4.2) -- (1.5,-1.8);
			
			\draw[gray] (-3.5,-3.5) -- (3.5,-3.5);
			
			\matrix[row sep=0.30cm,column sep=0.20cm] {
				\node (B1_k-1) [unitMS1] {$\mathbf{\mu}^\mathrm{q},\mathbf{\sigma}^\mathrm{q}$};&
				\node (a1_k-1) [unitA] {$\mathbf{a}_\mathrm{\mu},\mathbf{a}_\mathrm{\sigma}$};&
				\node (n1_k-1) [noise1] {$\mathbf{\epsilon}$};&
				\node (B1_k)   [unitMS1] {$\mathbf{\mu}^\mathrm{q},\mathbf{\sigma}^\mathrm{q}$};&
				\node (a1_k)   [unitA] {$\mathbf{a}_\mathrm{\mu},\mathbf{a}_\mathrm{\sigma}$};&
				\node (n1_k)   [noise1] {$\mathbf{\epsilon}$};&
				\node (B1_k+1) [unitMS1] {$\mathbf{\mu}^\mathrm{p},\mathbf{\sigma}^\mathrm{p}$};&
				\node (n1_k+1) [noise1] {$\mathbf{\epsilon}$};\\
				
				\node (A1_k-2) {};&
				\node (z1_k-1) [state1] {$\mathbf{z}$};&
				\node (A1_k-1) {};&
				\node (A11_k-1) {};&
				\node (z1_k)   [state1] {$\mathbf{z}$};&
				\node (A1_k)   {};&
				\node (A11_k) {};&
				\node (z1_k+1) [state1] {$\mathbf{z}$};&
				\node (A1_k+1) {};&
				\node (A11_k+1) {};\\
				
				\node (L1_k-2) {$\cdots$};&
				\node (D1_k-1) [unitD1] {$\mathbf{d}$};&
				\node (v11_k-1) {};&
				\node (v1_k){};&
				\node (D1_k) [unitD1] {$\mathbf{d}$};&
				\node (v11_k) {};&
				\node (v1_k+1) {};&
				\node (D1_k+1) [unitD1] {$\mathbf{d}$};&
				\node (v11_k+1) {};&
				\node (L1_k+2) {$\cdots$};\\
				\node (B0_k-1) [unitMS0] {$\mathbf{\mu}^\mathrm{q},\mathbf{\sigma}^\mathrm{q}$};&
				\node (a0_k-1) [unitA] {$\mathbf{a}_\mathrm{\mu}, \mathbf{a}_\mathrm{\sigma}$};&
				\node (n0_k-1) [noise0] {$\mathbf{\epsilon}$};&
				\node (B0_k)   [unitMS0] {$\mathbf{\mu}^\mathrm{q},\mathbf{\sigma}^\mathrm{q}$};&
				\node (a0_k) [unitA] {$\mathbf{a}_\mathrm{\mu}, \mathbf{a}_\mathrm{\sigma}$};&
				\node (n0_k)   [noise0] {$\mathbf{\epsilon}$};&
				\node (B0_k+1) [unitMS0] {$\mathbf{\mu}^\mathrm{p},\mathbf{\sigma}^\mathrm{p}$};&
				\node (n0_k+1) [noise0] {$\mathbf{\epsilon}$};\\
				\node (A0_k-2){};&
				\node (z0_k-1) [state0] {$\mathbf{z}$};&
				\node (A0_k-1) {};&
				\node (A00_k-1) {};&
				\node (z0_k) [state0] {$\mathbf{z}$};&
				\node (A0_k) {};&
				\node (A00_k) {};&
				\node (z0_k+1) [state0] {$\mathbf{z}$};&
				\node (A0_k+1) {};&
				\node (A00_k+1) {};\\
				\node (L0_k-2){$\cdots$};&
				\node (D0_k-1) [unitD0] {$\mathbf{d}$};&
				\node (v00_k-1) {};&
				\node (v0_k) {};&
				\node (D0_k) [unitD0] {$\mathbf{d}$};&
				\node (v00_k) {};&
				\node (v0_k+1) {};&
				\node (D0_k+1) [unitD0] {$\mathbf{d}$};&
				\node (v00_k+1) {};&
				\node (L0_k+2) {$\cdots$};\\ 
				\node (x_k-2) {};&
				\node (x_k-1) [prediction] {$\mathbf{x}$}; &
				\node (x0_k-1) {};&
				\node (x1_k-1) {};&
				\node (x_k) [prediction] {$\mathbf{x}$};&
				\node (x0_k) {};&
				\node (x1_k) {};&
				\node (x_k+1) [prediction] {$\mathbf{x}$}; &
				\node (x_k+2){}; \\
				
				\node (e_k-2) {};&
				\node (e_k-1) {$\mathbf{e}$};&
				\node (e0_k-2) {};&
				\node (e0_k-1) {};&
				\node (e_k) {$\mathbf{e}$};&
				\node (e0_k) {};&
				\node (e0_k+1) {};&
				\node (e_k+1) {};&
				\node (e0_k+2) {}; &
				\node (e_k+2) {};\\
				
				\node (y_k-2){};&
				\node (y_k-1) [sensation] {$\mathbf{y}$};&
				\node (x0_k-1){};&
				\node (x1_k-1){};&
				\node (y_k) [sensation] {$\mathbf{y}$};&
				\node (x0_k-1){};&
				\node (x1_k-1) {};&
				\node (y_k+1) {};&
				\node (y_k+2) {}; \\
				
				\node (t_k-2){Time\ \ $\cdots$};&
				\node (t_k-1) {$t-1$};&
				\node (t0_k-2){};&
				\node (t0_k-1){};&
				\node (t_k){$t$};&
				\node (t0_k){};&
				\node (t0_k+1){};&
				\node (t_k+1) {$t+1$};&
				\node (t0_k+2){};&
				\node (t_k+2) {$\cdots$};\\
			};
			
			\path[->]
			(z1_k-1) edge[color=gray] (D1_k-1)	
			(z1_k)   edge[color=gray] (D1_k)		
			(z1_k+1) edge[color=gray] (D1_k+1)
			
			(D1_k-1)   edge[color=gray, bend right=22] (D0_k)
			(D1_k)   edge[color=gray, bend right=22] (D0_k+1)
			(D0_k-1)   edge[color=gray, bend right=17] (D1_k)
			(D0_k)   edge[color=gray, bend right=25] (D1_k+1)
			
			(D1_k-1)   edge[color=gray, bend right=10] (B1_k)        
			(D1_k)   edge[color=gray, bend right=10] (B1_k+1)
			
			(L1_k-2) edge[color=gray] (D1_k-1)
			(D1_k-1) edge[color=gray] (D1_k)				
			(D1_k) edge[color=gray] (D1_k+1)				
			(D1_k+1) edge[color=gray] (L1_k+2)				
			
			(n1_k-1) edge[color=gray] (z1_k-1)				
			(n1_k)   edge[color=gray] (z1_k)
			(n1_k+1) edge[color=gray] (z1_k+1)
			
			(B1_k-1) edge[color=gray] (z1_k-1)
			(B1_k)   edge[color=gray] (z1_k)
			(B1_k+1) edge[color=gray] (z1_k+1)
			
			(a1_k-1) edge[color=gray] (B1_k-1)
			(a1_k)   edge[color=gray] (B1_k)

			(z0_k-1) edge[color=gray] (D0_k-1)	
			(z0_k)   edge[color=gray] (D0_k)		
			(z0_k+1) edge[color=gray] (D0_k+1)
			
			(D0_k)   edge[color=gray, bend right=10] (B0_k+1)
			(D0_k+1) edge[color=gray] (L0_k+2)		
			
			(L0_k-2) edge[color=gray] (D0_k-1)
			(D0_k-1) edge[color=gray] (D0_k)				
			(D0_k) edge[color=gray] (D0_k+1)				
			
			(n0_k-1) edge[color=gray] (z0_k-1)				
			(n0_k)   edge[color=gray] (z0_k)
			(n0_k+1) edge[color=gray] (z0_k+1)
			
			(B0_k-1) edge[color=gray] (z0_k-1)
			(B0_k)   edge[color=gray] (z0_k)
			(B0_k+1) edge[color=gray] (z0_k+1)
			
			(a0_k-1) edge[color=gray] (B0_k-1)
			(a0_k)   edge[color=gray] (B0_k)
			
			(D0_k-1) edge[color=gray] (B0_k)
			(D0_k-1) edge[color=gray] (x_k-1)
			(D0_k) edge[color=gray] (x_k)					
			(D0_k+1) edge[color=gray] (x_k+1)
			
			(x_k-1) edge[color=plotRed, densely dotted, bend right=15] (D0_k-1) 
			(x_k) edge[color=plotRed, densely dotted, bend right=15] (D0_k) 
			(D0_k-1) edge[color=plotRed, densely dotted, bend right=15] (z0_k-1) 
			(D0_k) edge[color=plotRed, densely dotted, bend right=15] (z0_k)
			(D0_k) edge[color=plotRed, densely dotted, bend right=10] (D0_k-1)
			(D0_k) edge[color=plotRed, densely dotted, bend left=17] (D1_k-1)
			(z0_k) edge[color=plotRed, densely dotted, bend right=10] (B0_k)
			(B0_k) edge[color=plotRed, densely dotted, bend right=5] (D0_k-1)
			(B0_k) edge[color=plotRed, densely dotted, bend right=15] (a0_k)               
			(z0_k-1) edge[color=plotRed, densely dotted, bend right=10] (B0_k-1)
			(B0_k-1) edge[color=plotRed, densely dotted, bend right=15] (a0_k-1)                
			
			(D1_k-1) edge[color=plotRed, densely dotted, bend right=15] (z1_k-1) 
			(D1_k) edge[color=plotRed, densely dotted, bend right=15] (z1_k)
			(D1_k) edge[color=plotRed, densely dotted, bend right=10] (D1_k-1)
			(D1_k) edge[color=plotRed, densely dotted, bend left=20] (D0_k-1)
			(z1_k) edge[color=plotRed, densely dotted, bend right=10] (B1_k)
			(B1_k) edge[color=plotRed, densely dotted, bend left=5] (D1_k-1)
			(B1_k) edge[color=plotRed, densely dotted, bend right=15] (a1_k)               
			(z1_k-1) edge[color=plotRed, densely dotted, bend right=10] (B1_k-1)
			(B1_k-1) edge[color=plotRed, densely dotted, bend right=15] (a1_k-1)  
			
			(x_k-1) edge[color=black] (e_k-1) 
			(x_k) edge[color=black] (e_k) 

			(e_k-1) edge[color=plotRed, densely dotted, bend right=15] (x_k-1) 
			(e_k) edge[color=plotRed, densely dotted, bend right=15] (x_k) 
			
			(y_k-1) edge[color=black] (e_k-1) 
			(y_k) edge[color=black] (e_k) 
			
			;
			
			\begin{pgfonlayer}{background}
			
			
			
			\end{pgfonlayer}
			\end{tikzpicture}
		\end{scriptsize}
		\caption{The notation has been simplified for clarity, so the $k$ and $t$ indexes have been dropped. The time constant $\iota^k$ (see Eq. \eqref{eq:d}) is greater in the High layer than in the Low layer. Since the High layer is the top on the hierarchy, the term $\mathbf{W}^{kk+1}_\mathrm{dd}\mathbf{d}^{k+1}_{t-1}$ is removed, analogously, the term $\mathbf{W}^{kk-1}_\mathrm{dd}\mathbf{d}^{k-1}_{t-1}$ is removed for the Low layer. The top-down process flow is represented by gray arrows. Red arrows illustrate the bottom-up process flow, where error is back propagated through time. The on-line inference process computed in a sliding time window is named Error Regression \cite{tani2003}.}
		\label{fig:VP-RNN}
	\end{figure}

	As a variational framework, PV-RNN includes a generative and an inference model (see Fig. \ref{fig:VP-RNN}). The generative model involves the prior distributions. It does hierarchical predictions on the output layer following a top-down flow. That is, the prediction is generated from the information represented in the latent state. The inference model involves posterior distributions, so the information flow goes in the bottom-up sense. Predictions are done starting from the evidence (i.e., a given observation), and the latent representation is modified in order to approximate the evidence.  
	
	Let\footnote{Notation: layer's latent states are denoted bold low-case, biases are denoted $\mathbf{b}$, weight connexion are denoted $\mathbf{W}$ with subscripts indicating the origin and destination of the connection (e.g., $\mathbf{W}_\mathrm{zd}$ are the weights connecting $\mathbf{z}$ to $\mathbf{d}$ units). Superscripts $k$ indicate the level in the MTRNN hierarchy. Finally, the superscripts p and q are used to distinguish between variables that belong to the prior and posterior distributions, respectively.} the generative model $P_\phi$ be defined from the parameters $\phi$, distributed among the components: generated prediction $\mathbf{x}$, stochastic $\mathbf{z}$ and the deterministic $\mathbf{d}$ latent states. For a prediction $\mathbf{x}_{1:T} = (\mathbf{x}_{1},\mathbf{x}_{2},..., \mathbf{x}_{T})$, and considering the parameters $\phi_\mathrm{x}$, $\phi_\mathrm{z}$, and $\phi_\mathrm{d}$, $P_\phi$ factorizes such that:
	
	\begin{equation}
	\begin{multlined}
	P_\phi\left(\mathbf{x}_{1:T},\mathbf{z}_{1:T},\mathbf{d}_{1:T}|\mathbf{z}_{0},\mathbf{d}_{0}\right) =\\
	\prod_{t=1}^{T}P_{\phi_\mathrm{x}}\left(\mathbf{x}_t |\mathbf{d}_t\right)P_{\phi_\mathrm{z}}\left(\mathbf{z}_t |\mathbf{d}_{t-1}\right)P_{\phi_\mathrm{d}}\left(\mathbf{d}_t |\mathbf{d}_{t-1}, \mathbf{z}_t\right)
	\end{multlined}
	\label{eq:prior}
	\end{equation}
	
	Let the deterministic states be defined according to a MTRNN structure. For the $k^\mathrm{th}$ context layer at time $t$, with timescale $\iota^k$, the internal dynamics are represented such that
	
	\begin{equation}
	\begin{array}{l}
	\mathbf{u}^{k}_{t} = \mathbf{W}^{kk}_\mathrm{dd}\mathbf{d}^k_{t-1} + \mathbf{W}^{kk-1}_\mathrm{dd}\mathbf{d}^{k-1}_{t-1} + \mathbf{W}^{kk+1}_\mathrm{dd}\mathbf{d}^{k+1}_{t-1} + \mathbf{W}^{kk}_\mathrm{dd}\mathbf{z}^k_{t} \\
	\mathbf{h}^{k}_{t} = \left(1-\frac{1}{\iota_k}\right)\mathbf{h}^k_{t-1} +\frac{1}{\iota_k}\mathbf{u}^{k}_{t}\\			
	\mathbf{d}^{k}_{t} = \mathrm{tanh}\left(\mathbf{h}^{k}_{t}\right)
	\end{array}.
	\label{eq:d}
	\end{equation} 
	
	The prior distribution $P_{\phi_\mathrm{z}}\left(\mathbf{z}_t|\mathbf{d}_{t-1}\right)$ is modeled as a Gaussian with diagonal covariance matrix, such that
	
	\begin{equation}
	P_{\phi_\mathrm{z}}\left(\mathbf{z}_t|\mathbf{d}_{t-1}\right) = \mathcal{N}\left(\mathbf{z}_t;\mathbf{\mu}^\mathrm{p}_t,\mathbf{\sigma}^\mathrm{p}_t\right),
	\label{eq:zp}
	\end{equation} 
	
	\noindent where $\mathbf{\mu}^\mathrm{p}_t$ and $\mathbf{\sigma}^\mathrm{p}_t$ are, respectively, the mean and standard deviation of $\mathbf{{z}_t} = \mathbf{\mu}^\mathrm{p}_{t}+\mathbf{\sigma}^\mathrm{p}_{t}\ast \mathbf{\epsilon}$, with $\mathbf{\epsilon}$ sampled from $\mathcal{N}(0,1)$. The variables $[\mathbf{\mu}^\mathrm{p}_t,\mathrm{log}\ ( \mathbf{\sigma}^\mathrm{p}_t)] = f_{\phi_{z}}(\mathbf{d}_{t-1})$ are obtained with $f_{\phi_{z}}(.)$ the one layer feed-forward neural network, such that 
	
	\begin{equation}
	\begin{array}{c}
	\mathbf{\mu}^{\mathrm{p},k}_t = \mathrm{tanh}\left(\mathbf{W}^{kk}_\mathrm{\mu d} \mathbf{d}^{k}_{t-1} + \mathbf{b}^{\mathrm{p},k}_\mathrm{\mu}\right)\\
	\mathrm{log}\left(\mathbf{\sigma}^{\mathrm{p},k}_t\right) = \mathbf{W}^{kk}_\mathrm{\sigma d} \mathbf{d}^{k}_{t-1} + \mathbf{b}^{\mathrm{p},k}_\mathrm{\sigma}
	\end{array}
	\label{eq:mup_lsp}.
	\end{equation}
	
	Let the inference model $Q_\pi$ (the approximate posterior) be defined from the parameters $\pi$, such that
	
	\begin{equation}
	Q_\pi(\mathbf{z}_t|\mathbf{d}_{t-1},\mathbf{e}_{t:T}) = \mathcal{N}\left(\mathbf{z}_t;\mathbf{\mu}^\mathrm{q}_t,\mathbf{\sigma}^\mathrm{q}_t\right),	
	\label{eq:posterior}
	\end{equation}
	
	\noindent where $\mathbf{\mu}^\mathrm{q}_t$ and $\mathbf{\sigma}^\mathrm{q}_t$ are, respectively, the mean and standard deviation of $\mathbf{{z}_t} = \mathbf{\mu}^\mathrm{q}_{t}+\mathbf{\sigma}^\mathrm{q}_{t}\ast \mathbf{\epsilon}$, with $\mathbf{\epsilon}$ sampled from $\mathcal{N}(0,1)$. The variables $[\mathbf{\mu}^\mathrm{q}_t,\mathrm{log}\ ( \mathbf{\sigma}^\mathrm{q}_t)] = f_{\pi_{z}}(\mathbf{d}_{t-1}, \mathbf{a}^\mathrm{\bar{x}})$ are obtained with $f_{\pi_{z}}(.)$ the one layer feed-forward neural network, such that 
	
	\begin{equation}
	\begin{array}{c}
	\mathbf{\mu}^{\mathrm{q},k}_t = \mathrm{tanh}\left(\mathbf{W}^{kk}_\mathrm{\mu d} \mathbf{d}^{k}_{t-1} + \mathbf{a}^{\mathrm{\bar{x}},k}_{\mathrm{\mu},t} + \mathbf{b}^{\mathrm{q},k}_\mathrm{\mu}\right)\\
	\mathrm{log}\left(\mathbf{\sigma}^{\mathrm{q},k}_t\right) = \mathbf{W}^{kk}_\mathrm{\sigma d} \mathbf{d}^{k}_{t-1} + \mathbf{a}^{\mathrm{\bar{x}},k}_{\mathrm{\sigma},t} + \mathbf{b}^{\mathrm{q},k}_\mathrm{\sigma}
	\end{array}.
	\label{eq:muq_lsq}
	\end{equation}
	
	The parameters $\mathbf{a}^\mathrm{\bar{x}}_{1:T}$ are introduced to provide the network with information about the prediction error in relation to a given pattern $\bar{\mathbf{x}}$. Thus, $\mathbf{a}^\mathrm{\bar{x}}_{1:T}$ is changed back propagating through time the prediction error $\mathbf{e}_{t:T}$, so information about the future steps of $\bar{\mathbf{x}}_{t:T}$, and existing dependencies with the current time step $t$, are captured. These terms are defined by
	
	\begin{equation}
	\begin{array}{c}
	\mathbf{a}^{\mathrm{\bar{x}},k}_{\mathrm{\mu},t} = \mathbf{a}^{\mathrm{\bar{x}},k}_{\mathrm{\mu},t} + \alpha\frac{\partial L}{\partial \mathbf{a}^{\mathrm{\bar{x}},k}_{\mathrm{\mu},t}}\\
	\mathbf{a}^{\mathrm{\bar{x}},k}_{\mathrm{\sigma},t} = \mathbf{a}^{\mathrm{\bar{x}},k}_{\mathrm{\sigma},t} + \alpha\frac{\partial L}{\partial \mathbf{a}^{\mathrm{\bar{x}},k}_{\mathrm{\sigma},t}}
	\end{array}.
	\label{eq:amu_als}
	\end{equation}
	
	\noindent with $\alpha$ denoting the learning rate. 
	
	Let the Variational Evidence Lower Bound (ELBO) $L(\phi,\pi)$ be defined by
	
	\begin{equation}
	\begin{multlined}
	L(\phi,\pi) = \sum_{t=1}^{T}\left( E_{\mathrm{q_\pi}}\left[\mathrm{log}\left(\mathbf{x}_t|\tilde{\mathbf{d}}_{t},\mathbf{z}_t\right)\right] -\right. \\
	\left.w\mathrm{KL}\left[Q_\mathrm{\pi}\left(\mathbf{z}_{t}|\tilde{\mathbf{d}}_{t-1},\mathbf{e}_{t:T}\right)\|P_{\phi_z}\left(\mathbf{z}_t|\tilde{\mathbf{d}}_{t-1}\right)\right]\right).
	\end{multlined}
	\label{eq:elbo}
	\end{equation}
	
	\noindent Since $\mathbf{d}_t$ is deterministic given $\mathbf{d}_{t-1}$ and $\mathbf{z}_t$, $\tilde{\mathbf{d}}_t$ denotes the center of a Dirac distribution. The first term at the right of the equation is a reconstruction component, it corresponds to the expected log-likelihood under the posterior distribution $Q_\mathrm{\pi}$. The second therm is a regulation component, it corresponds to the Kullback-Leibler (KL) divergence between the prior and the posterior distributions of the latent variables. The meta-parameter $w$ adjusts the optimization weight in learning the posterior and the prior distributions. After dropping the random variable notation to improve readability, the KL component can be expressed as

	\begin{equation}
	\begin{multlined}
	\mathrm{KL}\left[Q_\mathrm{\pi}\|P_{\phi_z}\right] =
	\mathrm{log}\left(\frac{\mathbf{\sigma}^\mathrm{p}}{\mathbf{\sigma}^\mathrm{q}}\right) + \frac{\left(\mathbf{\mu}^\mathrm{p}-\mathbf{\mu}^\mathrm{q}\right)^2 + \left(\mathbf{\sigma}^\mathrm{q}\right)^2}{2\left(\mathbf{\sigma}^\mathrm{p}\right)^2}-\frac{1}{2}
	\end{multlined}
	\label{eq:klDiv}
	\end{equation}
	
	Finally, the variable $\mathbf{x}_{i,t}$, related to the $i^{\mathrm{th}}$ dimension of the output space, is defined such that
	
	\begin{equation}
	\mathbf{x}_{i,t} = \mathrm{softmax}\left(\mathbf{W}_{\mathrm{dx}_i}\mathbf{d}^{0}_{t} + \mathbf{b}_{\mathrm{x}_i}\right).
	\label{eq:output}
	\end{equation}
	
	\noindent Unlike \cite{ahmadi2019}, we do not include in $\mathbf{x}_{i,t}$ connections from the stochastic latent distributions at the Low level.
	
	\section{MOTOR COMPLIANCE}
	\label{sec:control}

	In conformity with the principles of ethics in robotics experiments, inspired by the famous Asimov's three laws of robotics \cite{asimov1950}, motor compliance is studied within the context of social ethical conventions. Thus, the human and the robot actions must preserve each other's integrity and safety.
	
	Computational theories of human control have pointed out the plausibility of continuous and intermittent control systems in the brain, with the latter being possibly driven by events \cite{gawthrop2011}. Intermittent control is related to the prefrontal cortex, the premotor cortex, and the basal ganglia areas, it has been studied in the context of postural regulation \cite{tanabe2016}. 	
	
	We model motor control as a hybrid intermittent process driven by intentionality and social interaction. The rationale behind this is the following. When contact between the human and the robot is established, their intended motion may complement or be to some extent incongruent. Thus, in both the human and the robot cases, the body posture must be adapted to the other's influence to preserve a safe interaction. Expressed in other terms, once external forces acting on the body induce significant joint torques, the joint should exhibit viscoelasticity, so it becomes compliant to the external force. Once the external constraints cease to be relevant, active control should be resumed from a gradual transition to the desired state. Hence, in this hybrid view, each robot articulation is controllable instantaneously by one of two possible schemes: a compliant and an active scheme. 
	
	Let the switching between the compliant and the active modes of joint $j$ at time $t$ rely on a continuous observation process of the torque $\hat{\tau}_{j,t}$, and the prediction on the body dynamics $\tau^\mathrm{act}_{j,t}$. The estimation of the external torque induced in the interaction is such that
	
	\begin{equation}
	\hat{\tau}^{\mathrm{ext}}_{j,t} = \hat{\tau}_{j,t} - \tau^{\mathrm{act}}_{j,t}.
	\label{eq:extForce}
	\end{equation}
	
	\noindent Thus, switching to the compliant mode occurs once $|\hat{\tau}^{\mathrm{ext}}_{j,t}| > \tau^\mathrm{th}_{j}$ exceeds a threshold  $\tau^\mathrm{th}_{j}$.
	
	Let the joint position $\theta_{j,t}$ be regulated by the active control scheme in charge of tracking a reference position $\theta^\mathrm{net}_{j,t}$, intended by the agent (i.e. it is generated by the neural network). The target position $\theta^{act}_{j,t+1}$, in relation to the observation $\hat{\theta}_{j,t}$, is obtained from a discrete proportional control law, such that
	
	\begin{equation}
	\theta^{act}_{j,t+1} = \eta^\mathrm{a}_{j,t} \left(\theta^\mathrm{net}_{j,t} - \hat{\theta}_{j,t}\right).
	\label{eq:activeScheme}
	\end{equation}
	
	\noindent Since $\theta_{j,t}$ may differ considerably from $\theta^\mathrm{net}_{j,t}$ under the compliant scheme, in order to avoid abrupt motions, the proportional gain $\eta^\mathrm{a}_{j,t}$ is set as the cosine transition modulated by $\left|\theta^\mathrm{net}_{j,t} - \hat{\theta}_{j,t}\right|$, from a minimum to a maximum gain.  
	
	Let the joint position $\theta_{j,t}$ be regulated by the compliant control scheme in charge of following the torque induced by the human. A discrete time proportional integral (PI) feedback control law with gains $\eta^\mathrm{p}_j$ and $\eta^\mathrm{i}_j$ is adopted, so
	
	\begin{equation}
	\theta^\mathrm{ext}_{j,t+1} = \hat{\theta}_{j,t} + \eta^\mathrm{p}_j \hat{\tau}^{\mathrm{ext}}_{j,t} + f\left(\eta^\mathrm{i}_j\sum_t \hat{\tau}^\mathrm{ext}_{j,t}\right).
	\label{eq:extScheme}
	\end{equation}
	
	\noindent In both the active and the compliant schemes the target correction is saturated to preserve safety in behavior. Thus, the function $f(.)$ acts as a reset windup to the integral term.
	
	As explained before, due to the fact that the control is done in the joint space, an interesting situation emerges where the joints may be set to different control schemes. Hence, the human provides feedback to the robot through the compliant joints, while receiving feedback about the robot's intention from the active joints. However, it is also interesting to include soft impedance in the compliant joints in order to enhance the interaction experience. Thus, the target position in the compliant scheme is obtained such that
	
	\begin{equation}
	\theta^{com}_{j,t+1} = \theta^\mathrm{ext}_{j,t+1} + \eta^\mathrm{n}_j s\left(\theta^\mathrm{net}_{j,t} - \hat{\theta}_{j,t}\right), 
	\label{eq:compliantScheme}
	\end{equation}
	
	\noindent with $\eta^\mathrm{n}_j$ a gain parameter and $s(.)$ a saturation function.	
	
	\section{METHODOLOGY}
	\label{sec:methodology}
	
	On the robot side, the variables studied were \textit{intentionality} (desired behavior in relation to the human's actions) and \textit{compliance} in  the physical (or motor, M) and the representational (or cognitive, C) dimensions. As shown in Table \ref{tab:complicance}, several profiles were defined to study compliance. Regarding the motor dimension, the exclusively compliant mode $\mathrm{M_{teach}}$ was employed for kinesthetic demonstration of the behavior primitives, and the interaction mode $\mathrm{M_{inter}}$ was set based on the hybrid controller (including the active and the compliant schemes). Concerning the cognitive dimension, the profiles $\mathrm{C_{rigid}}$, $\mathrm{C_{mod}}$, and $\mathrm{C_{flex}}$ were defined corresponding, respectively, to a rigid, a moderate, and a flexible agent.

	\begin{table}[th]
		\caption{The cognitive and motor profiles.}
		\begin{center}
			\begin{tabular}{l p{6.5cm}}
				\textbf{Profile} & \textbf{Description}\\
				\hline
				$\mathrm{M_{teach}}$ & Kinesthetic teaching (with the compliance scheme only)\\
				$\mathrm{M_{inter}}$& Interaction (with the active and compliant schemes)\\
				$\mathrm{C_{rigid}}$ & Strong intentionality ($w=0.01$)\\
				$\mathrm{C_{mod}}$ & Moderate intentionality ($w=0.001$)\\
				$\mathrm{C_{flex}}$& Low intentionality ($w=0.0001$)\\
				\hline		
			\end{tabular}
		\end{center}		
		\label{tab:complicance}
	\end{table}
	
\begin{figure}[thpb]
		\centering
		\begin{scriptsize}			
			\begin{tikzpicture}
			\node [] at (0,0){
				\includegraphics[width=0.47\textwidth,keepaspectratio]{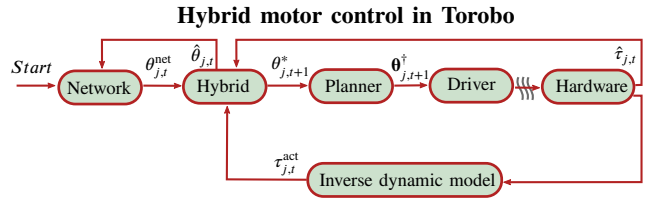}};
			\node [color=black] at (0.2,1.35) {\small \textbf{Hybrid motor control in Torobo}};				
			\begin{scope}[shift={(0.02,0.02)}]
			\node at (-3.1,0.4) {Network};
			\node at (-1.45,0.4) {Hybrid};
			\node at (0.28,0.42) {Planner};
			\node at (1.85,0.45) {Driver};
			\node at (3.45,0.42) {Hardware};
			\node at (1.0,-0.85) {Inverse dynamic model};
			\begin{scriptsize}			
			\node at (-4.0,0.67) {$Start$};			
			\node at (-2.3,0.67) {$\theta^\mathrm{net}_{j,t}$};
			\node at (-1.72,0.82) {$\hat{\theta}_{j,t}$};			
			\node at (-0.55,0.67) {$\theta^\ast_{j,t+1}$};			
			\node at (1.05,0.67) {$\mathbf{\theta}^{\dagger}_{j,t+1}$};
			\node at (3.9,0.82) {$\hat{\tau}_{j,t}$};			
			\node at (-0.6,-0.62) {$\tau^\mathrm{act}_{j,t}$};			
			
			\end{scriptsize}			
			\end{scope}
			\end{tikzpicture}
		\end{scriptsize}
		\caption{The Network block includes the PV-RNN model, it computes the desired position $\theta^\mathrm{net}_{j,t}$ based on the current estimation $\hat{\theta}_{j,t}$. From measured $\hat{\tau}_{j,t}$ and estimated torque $\tau^\mathrm{act}_{j,t}$ (by inverse dynamics), the Hybrid controller computes the next target $\theta^\ast_{j,t+1}$. The Planner calculates a trajectory for the target in open-loop via intermediate positions $\mathbf{\theta}^{\dagger}_{j,t+1}$. Lately, the robot Driver manages the Hardware plant.}
		\label{fig:control}
	\end{figure}

\begin{figure}[h!]
		\centering		
		\begin{tabular}{c}		
			\begin{tikzpicture}			
			\node [] at (0,0){
				\includegraphics[width=0.47\textwidth,keepaspectratio]{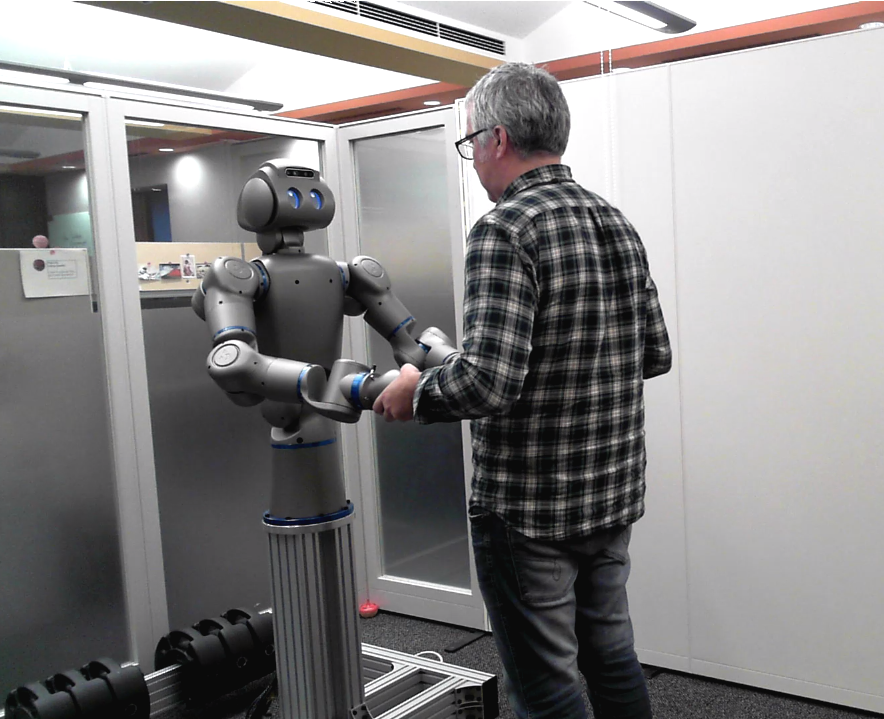}};
			\node [color=black] at (-0.2,3.75) {\small \textbf{The experimental and simulation environments}};				
			\end{tikzpicture}\\
			\begin{tikzpicture}			
			\node [] at (0,0){
				\includegraphics[width=0.47\textwidth,keepaspectratio]{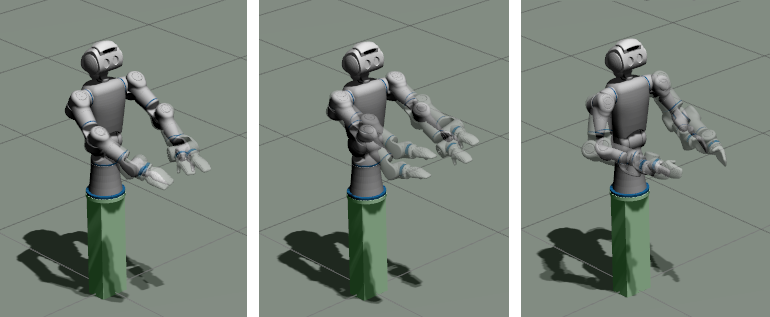}};				
			\path[-stealth,line width=0.7pt,densely dotted]
			
			(-2.3,-0.5) edge[color=orange] (-1.7,-0.2)	
			(-1.7,-0.2) edge[color=orange] (-2.3,-0.5)	
			
			(0.6,0.5) edge[color=orange] (0.6,-0.5)	
			(0.6,-0.5) edge[color=orange] (0.6,0.5)	
			
			(3.0,-0.4) edge[color=orange] (2.3,-0.2)	
			(2.3,-0.2) edge[color=orange] (3.0,-0.4)	
			
			;
			
			\node [color=white] at (-1.9,-1.5) {\scriptsize \textbf{A}};
			\node [color=white] at (1.0,-1.5) {\scriptsize \textbf{B}};
			\node [color=white] at (3.8,-1.5) {\scriptsize \textbf{C}};			
			\end{tikzpicture} 
		\end{tabular}					
		\caption{The primitives visualized in the Gazebo simulator were captured by kinesthetic demonstration with Torobo.}
		\label{fig:primitives}		
	\end{figure}

	The implementation of hybrid motor control in Torobo is illustrated in Fig. \ref{fig:control}. A dataset was constituted with three behavior primitives (see Fig. \ref{fig:primitives}), sampled at 4 Hz, during 90 time steps. The cognitive profiles in Table \ref{tab:complicance} were modeled in the PV-RNN architecture (see Fig. \ref{fig:VP-RNN}). It included a Low layer (40 d units, 4 z units, and $\iota = 2$) and a High layer (10 d units, 1 z unit, and $\iota = 10$). All learnable variables  were updated during the training phase, whereas in interaction mode only the terms $\mathbf{a}$ were updated (see Eqs. \eqref{eq:muq_lsq}\eqref{eq:amu_als}). Conforming to Table \ref{tab:complicance}, specific values for $w$ (Eq. \eqref{eq:elbo}) were selected for training (50K epochs). For interaction $w=1.0\mathrm{e}-5$ was set for all the profiles, since it produced the best results. 
	
	On the human side, the variables under study are \textit{intentionality} (desired behavior in relation to the robot's actions) and \textit{engagement} (physical efforts invested in the task). Finally, on the mutual interaction side, the variable \textit{behavior emergence} (how similar the robot's posture is to the known primitives) is studied through a regression observer. Since the motion primitives were reasonably different from one another, a feed-forward model was designed (12 $\mathrm{Input}$, 150 $\mathrm{Hidden}_1$, 15 $\mathrm{Hidden}_2$, 3 $\mathrm{Output}$, with tanh activation for the hidden layers and sigmoid activation for the output layer), and trained by supervised learning with the joint instantaneous positions. A success rate of 100\% was achieved in the test set.
	
	\paragraph*{Experimental protocol} Six scenarios were portrayed conforming the pairs AB, AC, BA, BC, CA, CB; with the left term representing the robot's intended behavior, and the right term corresponding to the human intended behavior (the subject was instructed to induce a given primitive in the robot behavior). For example, AB means the robot wants to do A and the human wants to do B. A total of 54 trials were registered for the experimental subject (6 pairs x 3 cognitive models x 3 times), during 300 time steps each.
	
	\paragraph*{Software platform}

The open-source implementation of the models is provided by the \textit{neural robotics library} \cite{chame2020}. From previous experiences \cite{chame2016}, C++ was chosen as the base programming language. The programs run in the Robot Operative System (ROS) Kinetic Kane over Ubuntu 16.04 LTS. The Network block (Fig. \ref{fig:control}) ran at 4 Hz in the host computer (Alienware Aurora R7, 12 Intel\textregistered \ Core\texttrademark \ i7-8700K CPU at 3.70GHz, and 31.1 GiB RAM memory). In interaction mode, BPTT was computed within a sliding window (20 time steps) during 28 epochs. The models learned 12 degrees of freedom (6 for each arm), constant desired references were given to the torso and the head joints.
  
	
	\section{RESULTS}
	\label{sec:results}
\begin{figure}[thpb]			
		\begin{tikzpicture}
		\node [] at (0,0){
			\includegraphics[width=0.47\textwidth,keepaspectratio]{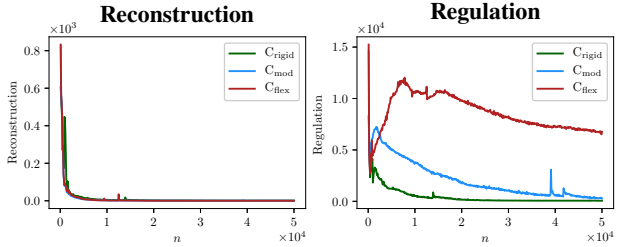}};			
				\node [color=black] at (-1.8,1.6) {\small \textbf{Reconstruction}};				
				\node [color=black] at (2.3,1.6) {\small \textbf{Regulation}};				
							
		\end{tikzpicture}
		\caption{Training during 50k epochs. The reconstruction and the regulation terms of the ELBO (see Eq. \eqref{eq:elbo}).}			
		\label{fig:training}
	\end{figure}

	The training results are shown in Fig. \ref{fig:training}. As noticed, although the reconstruction component of the ELBO (see Eq. \eqref{eq:elbo}) approached to zero in all cases, which indicates good predictions from the posterior distribution; the smaller the meta-parameter $w$ was set (right plot), the more dissimilar the posterior and the prior distributions were, which implies more stochasticity in the generative process.
	
	\subsection{Simulations}
	\label{sec:simulation}
\begin{figure}[h]			
		\begin{tikzpicture}
		\node [] at (0,0){			\includegraphics[width=0.47\textwidth,keepaspectratio]{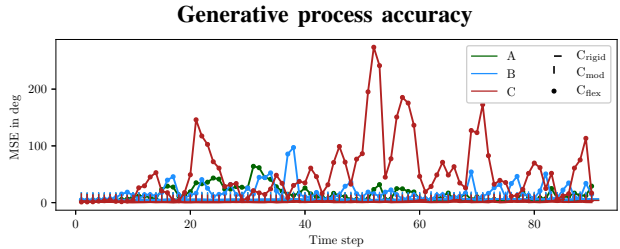}};				
		\node [color=black] at (0.1,1.65) {\small \textbf{Generative process accuracy}};	
					
		\end{tikzpicture}
		\caption{Comparison between the references constituting the training dataset and the learned generation.}			
		\label{fig:MSEPred}
	\end{figure}

	Two preliminary studies were conducted. The first one investigated the accuracy of the generative process. For this, the primitives were generated during 90 time steps (the same length of the captured sequences in the dataset) by each cognitive model (see Table \ref{tab:complicance}). The comparison is done by calculating at each time step the mean squared error (MSE) between the reference $\bar{\mathbf{x}}$ and the generated $\mathbf{x}$ sequences. The models $\mathrm{C_{rigid}}$ and $\mathrm{C_{mod}}$ performed similarly well (see Fig. \ref{fig:MSEPred}), whereas $\mathrm{C_{flex}}$ had a more stochastic generative process.
	
\begin{figure}[h]			
		\begin{tikzpicture}
		\node [] at (0,0){
			\includegraphics[width=0.47\textwidth,keepaspectratio]{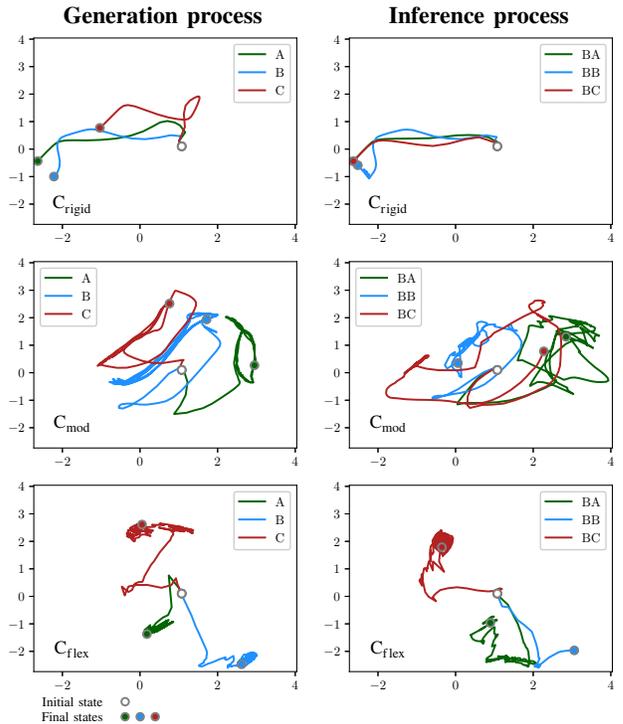}};
		\node [color=black] at (2.2,4.6) {\small \textbf{Inference process}};				
		\node [color=black] at (-2.0,4.6) {\small \textbf{Generation process}};				
		\node [color=black] at (-3.2,2.1) {\scriptsize $\mathrm{C_{rigid}}$};
		\node [color=black] at (-3.2,-0.8) {\scriptsize $\mathrm{C_{mod}}$};
		\node [color=black] at (-3.2,-3.8) {\scriptsize $\mathrm{C_{flex}}$};
		\node [color=black] at (1.0,2.1) {\scriptsize $\mathrm{C_{rigid}}$};
		\node [color=black] at (1.0,-0.8) {\scriptsize $\mathrm{C_{mod}}$};
		\node [color=black] at (1.0,-3.8) {\scriptsize $\mathrm{C_{flex}}$};
		\begin{scope}[shift={(-3.2,-4.7)}]
			\node [color=black] at (0.0,0.2) {\tiny Initial state};
			\path [draw=gray,fill=white,line width=0.25mm] (0.7,0.2) circle (0.5mm);
			\node [color=black] at (0.0,0.0) {\tiny Final states};
			\path [draw=gray,fill=plotGreen,line width=0.25mm] (0.7,0) circle (0.5mm);
			\path [draw=gray,fill=plotBlue,line width=0.25mm] (0.9,0) circle (0.5mm);
			\path [draw=gray,fill=plotRed,line width=0.25mm] (1.1,0) circle (0.5mm);
		\end{scope}
		
		\end{tikzpicture}		
		\caption{Two-PCA High layer's states $\mathbf{d}$. Agents were set to generate B (intention), but received A, B, and C as evidence.}
		\label{fig:simPrim}
	\end{figure}

	The second study analyzed cognitive compliance through the latent state $\mathbf{d}$. Figure \ref{fig:simPrim} presents a comparison on two principal component analysis (PCA) for the generation process of the primitives A, B, and C (left column), and the inference process (right column). Errors were BPTT by taking the difference between the model's prediction and recorded joint positions of A, B, and C. What is being analyzed is the possibility of transition from the generation of one primitive to another, given the evidence. That is, to what extent stochasticity from the hidden random distributions, received through the parameters $\mathbf{a}$ (see Eqs. \eqref{eq:muq_lsq}\eqref{eq:amu_als}), is allowed to affect the contextual representation during on-line inference. All transitions could be obtained, except from BC with $\mathrm{C_{rigid}}$. As noticed, $\mathrm{C_{flex}}$ representations were simpler, and converged faster to conform the evidence.  
	
	\subsection{Experiments}
	\label{sec:experiments}
	
\begin{figure}[h]			
		\begin{tikzpicture}
		\node [] at (0,0){
			\includegraphics[width=0.47\textwidth,keepaspectratio]{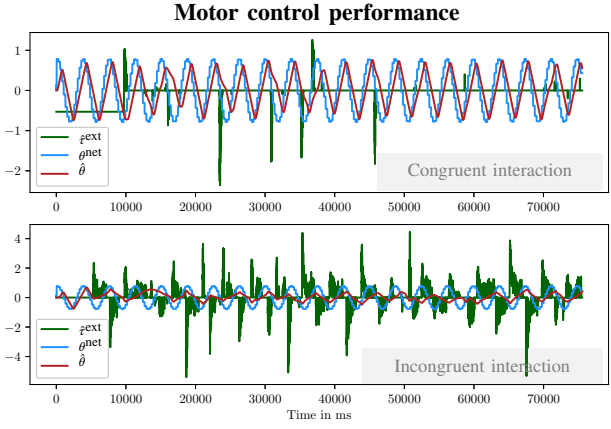}};
		\node [color=black] at (0.1,2.8) {\small \textbf{Motor control performance}};
		
		\node [color=gray] at (2.4,0.7) {\scriptsize Congruent interaction};
		\node [color=gray] at (2.3,-1.9) {\scriptsize Incongruent interaction};
		\node (rect) at (2.4,0.7) [fill opacity=0.1, draw opacity=0.0, draw=white,fill=gray, minimum width=3.0cm,minimum height=0.5cm] {};
		\node (rect) at (2.3,-1.9) [fill opacity=0.1, draw opacity=0.0, draw=white,fill=gray, minimum width=3.2cm,minimum height=0.5cm] {};
		\draw [fill=white,white] (-3.1,1.2) rectangle (-2.75,0.6);
		\node at (-2.9,1.15) {\tiny {$\hat{\tau}^\mathrm{ext}$}};
		\node at (-2.9,0.94) {\tiny {$\theta^\mathrm{net}$}};
		\node at (-3.02,0.74) {\tiny {$\hat{\theta}$}};
		\begin{scope}[shift={(0,-2.54)}]
		\draw [fill=white,white] (-3.1,1.2) rectangle (-2.75,0.6);
		\node at (-2.9,1.15) {\tiny {$\hat{\tau}^\mathrm{ext}$}};
		\node at (-2.9,0.94) {\tiny {$\theta^\mathrm{net}$}};
		\node at (-3.02,0.74) {\tiny {$\hat{\theta}$}};
		\end{scope}
		
		\end{tikzpicture}
		\caption{Torobo's right elbow follows a sinusoidal limit cycle reference. Torque in Nm, angle in rad.}
		\label{fig:contolStudy}	
	\end{figure}

\begin{figure}[h]			
		\begin{tikzpicture}
		\node [] at (0,0){
			\includegraphics[width=0.47\textwidth,keepaspectratio]{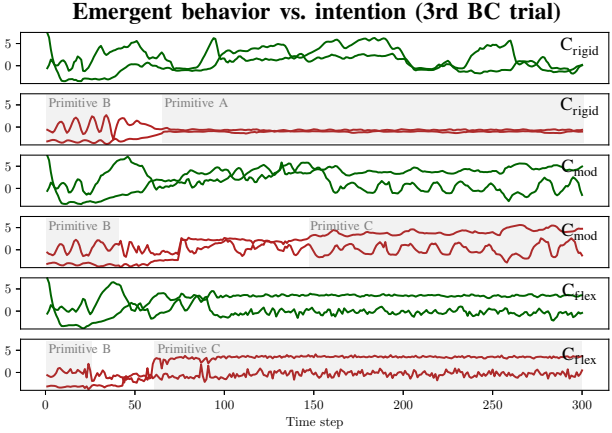}};				
		
		\node [color=black] at (0.1,2.85) {\small \textbf{Emergent behavior vs. intention (3rd BC trial)}};
		\node (rect) at (-3.07,1.45) [fill opacity=0.1, draw opacity=0.0, draw=white,fill=gray, minimum width=0.85cm,minimum height=0.65cm] {};
		\node (rect) at (0.85,1.45) [fill opacity=0.1, draw opacity=0.0, draw=white,fill=gray,minimum width=5.6cm,minimum height=0.65cm] {};
		\node at (-3.05,1.67) [color=gray] {\tiny Primitive B};
		\node at (-1.5,1.67) [color=gray] {\tiny Primitive A};
		\begin{scope}[shift={(0.0,-1.64)}]
		\node (rect) at (-3.015,1.45) [fill opacity=0.1, draw opacity=0.0, draw=white,fill=gray, minimum width=0.96cm,minimum height=0.65cm] {};
		\node (rect) at (1.8,1.45) [fill opacity=0.1, draw opacity=0.0, draw=white,fill=gray,minimum width=3.6cm,minimum height=0.65cm] {};
		\node at (-3.05,1.67) [color=gray] {\tiny Primitive B};
		\node at (0.43,1.67) [color=gray] {\tiny Primitive C};
		\end{scope}
		\begin{scope}[shift={(0.0,-3.28)}]
		\node (rect) at (-3.2,1.45) [fill opacity=0.1, draw opacity=0.0, draw=white,fill=gray,minimum width=0.6cm,minimum height=0.65cm] {};
		\node (rect) at (0.78,1.45) [fill opacity=0.1, draw opacity=0.0, draw=white,fill=gray,minimum width=5.68cm,minimum height=0.65cm] {};
		\node at (-3.05,1.67) [color=gray] {\tiny Primitive B};
		\node at (-1.6,1.67) [color=gray] {\tiny Primitive C};
		\end{scope}
		
		\node at (3.6,1.6) {\scriptsize $\mathrm{C_{rigid}}$};
		\node at (3.6,2.4) {\scriptsize $\mathrm{C_{rigid}}$};
		\begin{scope}[shift={(0.0, -1.55)}]			
		\node at (3.6,1.5) {\scriptsize $\mathrm{C_{mod}}$};
		\node at (3.6,2.35) {\scriptsize $\mathrm{C_{mod}}$};				
		\end{scope}
		\begin{scope}[shift={(0.0, -3.2)}]			
		\node at (3.6,1.5) {\scriptsize $\mathrm{C_{flex}}$};
		\node at (3.6,2.35) {\scriptsize $\mathrm{C_{flex}}$};
		\end{scope}
		\end{tikzpicture}
		\caption{Two-PCA, emergent behavior: green, intention: red.}			
		\label{fig:BvsI}
	\end{figure}

	
	The performance of the intermittent controller for congruent and incongruent interaction is shown in Fig. \ref{fig:contolStudy}. As noticed, smooth trajectories resulted from tracking the network signal while complying to the external torque induced by the human. Table \ref{tab:probTrans} compares emergent behavior to the robot's intended behavior for the three cognitive agents, based on the regression observer's evaluation. Table \ref{tab:effort} shows the mean and standard deviation of the estimated external torque during the experiment. In Fig. \ref{fig:BvsI}, emergent behavior for the pair BC is compared to the intended behavior of the robot\footnote{The experiment video is available at: https://youtu.be/f4TXmB7HV-s}.
	
	\begin{table}[th]
		\caption{Probability of the robot intending $I$ and behaving $B$, according to the desired motions induced by the human.}
		\begin{center}
			\begin{tabular}{l c c c c c c}
				& \multicolumn{2}{c}{$\mathbf{C_\mathrm{rigid}}$} & \multicolumn{2}{c}{$\mathbf{C_\mathrm{mod}}$} & \multicolumn{2}{c}{$\mathbf{C_\mathrm{flex}}$}\\
				\cline{2-7}
				\textbf{Data} & $p(I)$ & $p(B)$ & $p(I)$ & $p(B)$ & $p(I)$ & $p(B)$\\\hline
				
				Simulation	& 0.466 & 0.467 & 0.761 & 0.759 & \textbf{0.843} & \textbf{0.841}\\
				Exp. trial 1 	& 0.122 & 0.365 & \textbf{0.562} & \textbf{0.588} & 0.245 & 0.393\\
				Exp. trial 2 	& 0.274 & 0.476 & \textbf{0.705} & \textbf{0.686} & 0.394 & 0.451\\
				Exp. trial 3 	& 0.302 & 0.498 & \textbf{0.590} & \textbf{0.619} & 0.405 & 0.501\\
				
				\hline												
			\end{tabular}
		\end{center}		
		\label{tab:probTrans}
	\end{table}
	
	\begin{table}[th]
		\caption{External torque $\sum_j |\hat{\tau}^\mathrm{ext}_{j,t}|$ in Nm for trials.}
		\begin{center}			
			\begin{tabular}{c c c c c c c c c c}				
				& \multicolumn{3}{c}{$\mathbf{C_\mathrm{rigid}}$} & \multicolumn{3}{c}{$\mathbf{C_\mathrm{mod}}$} & \multicolumn{3}{c}{$\mathbf{C_\mathrm{flex}}$}\\
				\cline{2-10}				
				Case & $T_1$ & $T_2$ & $T_3$ & $T_1$ & $T_2$ & $T_3$ & $T_1$ & $T_2$ & $T_3$ \\
				\hline
				\multicolumn{10}{c}{Mean}\\ \hline
				AB & 2.0 & \textbf{1.4} & 1.9 & \textbf{1.9} & 1.8 & \textbf{1.8} & 2.0 & 2.0 & 1.9\\
				AC & 2.4 & 2.0 & \textbf{1.1} & \textbf{2.2} & \textbf{1.8} & 1.7 & 2.4 & 2.0 & 2.2\\
				BA & 1.9 & 1.6 & \textbf{1.4} & \textbf{1.6} & \textbf{1.4} & 1.6 & 1.7 & \textbf{1.4} & \textbf{1.4}\\
				BC & 2.2 & 2.0 & 1.4 & \textbf{2.1} & 1.9 & 2.0 & 2.3 & \textbf{1.8} & 2.2\\
				CA & \textbf{1.9} & 2.2 & \textbf{1.9} & 2.0 & \textbf{1.5} & 2.1 & 1.8 & 1.9 & 2.2\\
				CB & \textbf{1.9} & 2.1 & \textbf{1.8} & \textbf{1.9} & \textbf{1.9} & \textbf{1.8} & 2.0 & 2.2 & 2.1\\
				
				\hline
				\multicolumn{10}{c}{Standard Deviation}\\ \hline
				AB & 0.9 & 0.8 & 0.9 & 1.0 & 1.0 & 0.9 & 1.1 & 1.0 & 1.0\\
				AC & 1.2 & 0.9 & 0.7 & 1.0 & 0.9 & 0.8 & 1.1 & 1.0 & 1.1\\
				BA & 1.0 & 0.9 & 0.7 & 0.9 & 0.9 & 0.8 & 0.9 & 0.8 & 0.8\\
				BC & 1.0 & 0.9 & 0.8 & 1.0 & 0.9 & 1.0 & 1.2 & 0.9 & 1.0\\
				CA & 0.9 & 0.9 & 0.8 & 1.0 & 1.0 & 1.3 & 0.9 & 0.9 & 1.0\\ 
				CB & 0.8 & 1.0 & 0.8 & 0.9 & 1.0 & 0.8 & 1.0 & 1.1 & 0.9\\
				
				\hline
			\end{tabular}
		\end{center}
		\label{tab:effort}
	\end{table}
	
	\subsection{Discussion}
	\label{sec:discussion}
	
	The results supported the plausibility to model distinct cognitive styles through the meta parameter $w$. Thus, by learning to strongly approximate the prior and posterior distributions (see Fig \ref{fig:training}), the rigid agent was able to generate accurate behavior (see Fig \ref{fig:MSEPred}), but it was less sensitive to evidence. Indeed, it obtained greater differences between the intended and observed behaviors (see Table \ref{tab:probTrans}). Contrarily, the flexible agent was trained investing less efforts in learning to approximate these distributions, which resulted in stochastic or hesitating behavior. The moderate agent presented a good balance between accuracy and cognitive flexibility.
	
	
	When analyzing the human engagement in the interaction, qualitative differences can be observed among the agents (see Table \ref{tab:effort}, the lowest mean values per trial are highlighted in bold). Since the subject was instructed to induce as long as possible a certain behavior, with the rigid agent efforts were probably more invested in modifying the robot posture to encourage the generation of the desired primitive, given the agent's reluctance to change. Hence, the interactions were arduous and relied mostly on the compliant component of the hybrid motor control scheme. Considering the flexible agent, it likely adopted the desired posture shape, but efforts were required to keep consistent interaction. It was challenging to induce gradual changes in behavior, due to erraticness and loss of bilateral symmetry, which would explain the disparity of simulated and experimental results. In relation to the moderate agent, smoother interactions were obtained and less efforts appeared to be invested, since the agent was able to both change the posture and generate the behavior consistently. Finally, in agreement with \cite{Ikemoto12}, learning probably occurred on the human side, since both $p(I)$ and $p(B)$ were generally larger from the first to the third trials.
	

	\section{CONCLUSIONS AND FUTURE WORK}
	\label{sec:conclusion}
	

	This work focused on the study of physical interaction between a human and a robot, and considered both coherent and incoherent scenarios. An important distinction was established between motor and cognitive compliance. A variational model, inspired by the principles of predictive coding and active inference, was proposed to model cognitive compliance as the capacity to be driven by sensory evidence. An intermittent control concept was proposed to study motor deliberation while adapting to the human interaction, based on torque feed-back. The experiments results pointed out a trade-off between cognitive compliance and refinement in autonomous motion. We believe that this trade-off can be explored in developmental robotics to investigate on-line learning \cite{Ikemoto12}. From the perspective of human-robot interaction research, our results would also open interesting possibilities for the study of social cognition in human science, including topics in \textit{motivation} \cite{chame2019} and \textit{intersubjectivity} \cite{chame2020}.
	
	\addtolength{\textheight}{-12.3cm}   
	


	
	
	
	
	\bibliographystyle{IEEEtran}
	\bibliography{references}
	
\end{document}